\documentclass[journal]{IEEEtran}
\ifCLASSINFOpdf
   \usepackage[pdftex]{graphicx}
   \graphicspath{{figures/}}
\else
\fi
\usepackage{amsmath,amsfonts,amssymb,amsthm}
\DeclareMathOperator*{\minimize}{minimize}
\usepackage{multicol}
\usepackage{xcolor}
\usepackage[textsize=tiny]{todonotes}

\newcommand{\norm}[1]{\lVert #1 \rVert}
\usepackage{multirow}
\usepackage{hyperref}

\usepackage{algorithmicx, algpseudocode, algorithm}
\usepackage{amssymb}

\usepackage{cite}

\ifCLASSOPTIONcompsoc
 \usepackage[caption=false,font=normalsize,labelfont=sf,textfont=sf]{subfig}
\else
 \usepackage[caption=false,font=footnotesize]{subfig}
\fi
\begin{document}
%
\title{Dynamic Precision Analog Computing for Neural Networks}
%
%
%

\author{Sahaj~Garg,
        Joe~Lou,
        Anirudh~Jain,
        and~Mitchell~Nahmias
\thanks{S. Garg, J. Lou, A. Jain, and M. Nahmias are affiliated with Luminous Computing, Mountain View, CA 94305 USA. (e-mail: sahajgarg@gmail.com, zlou@alumni.stanford.edu, anirudh1097@gmail.com, mitch@luminouscomputing.com). Preprint.}}

\maketitle

\newcommand{\sahaj}[1]{{\color{blue} [Sahaj: #1]}}
\newcommand{\kate}[1]{{\color{blue} [Kate: #1]}}
\newcommand{\matt}[1]{{\color{black!50!green} [Matt: #1]}}
\newcommand{\rohun}[1]{{\color{orange} [Rohun: #1]}}
\newcommand{\rodolfo}[1]{{\color{orange} [Rodolfo: #1]}}
\newcommand{\joe}[1]{{\color{black!50!green} [Joe: #1]}}
\newcommand{\mitch}[1]{{\color{red} [Mitch: #1]}}
\newcommand{\anirudh}[1]{{\color{red} [Anirudh: #1]}}

\begin{abstract}
Analog electronic and optical computing exhibit tremendous advantages over digital computing for accelerating deep learning when operations are executed at low precision.
In this work, we derive a relationship between analog precision, which is limited by noise, and digital bit precision. 
We propose extending analog computing architectures to support varying levels of precision by repeating operations and averaging the result, decreasing the impact of noise. Such architectures enable programmable tradeoffs between precision and other desirable performance metrics such as energy efficiency or throughput. 
To utilize dynamic precision, we propose a method for learning the precision of each layer of a pre-trained model without retraining network weights.
We evaluate this method on analog architectures subject to a variety of noise sources such as shot noise, thermal noise, and weight noise and find that employing dynamic precision reduces energy consumption by up to 89\% for computer vision models such as Resnet50 and by 24\% for natural language processing models such as BERT. In one example, we apply dynamic precision to a shot-noise limited homodyne optical neural network and simulate inference at an optical energy consumption of 2.7 aJ/MAC for Resnet50 and 1.6 aJ/MAC for BERT with ${<}2\%$ accuracy degradation.

\end{abstract}


%
\IEEEpeerreviewmaketitle

\newcommand{\bfx}{\mathbf{x}}
\newcommand{\bfw}{\mathbf{w}}
\newcommand{\bfW}{\mathbf{W}}
\newcommand{\bfX}{\mathbf{X}}
\newcommand{\bfA}{\mathbf{A}}
\newcommand{\bfa}{\mathbf{a}}
\newcommand{\bfE}{\mathbf{E}}
\newcommand{\bftheta}{\mathbf{\theta}}
\newcommand{\xreal}{\bfx}
\newcommand{\wreal}{\bfw}
\newcommand{\xq}{\bfx^{(q)}}
\newcommand{\wq}{\bfw^{(q)}}
\newcommand{\yq}{y^{(q)}}
\newcommand{\zr}{z}
\newcommand{\R}{\mathbb{R}}
\newcommand{\N}{\mathcal{N}}
\newcommand{\ytilde}{\tilde{y}}
\newcommand{\ytq}{\tilde{y}^{(q)}}
\newcommand{\ztr}{\tilde{z}}

\section{Introduction}
\IEEEPARstart{A}{nalog} electronic and optical computing have demonstrated significant promise for accelerating matrix multiplications, the dominant computational cost in deep learning \cite{Nahmias:MAC}. Common approaches include resistive crossbar arrays \cite{fick:mythic, Joshi2020AccurateDN, isaac} and passive linear optical circuits  \cite{deLima:neuromorpic-ml,hamerly:homodyne,shen:coherent,Tait:Broadcast}. 
By minimizing data movement costs (either using in-memory processing or moving data optically), amortizing energy over a large matrix multiplication, and completing entire matrix multiplications in a single clock cycle, analog processors can exhibit improvements over digital electronics in energy (${>}10^2$), speed (${>}10^3$), and compute density (${>}10^2$) \cite{Nahmias:MAC}. 
These performance improvements are critical as deep learning models double in size every 3.4 months  \cite{amodei:openai}, outpacing the growth of Moore's law and stretching the limits of digital hardware.
However, analog computing exhibits these tremendous advantages over digital computing only when operations can be executed at low precision 
\cite{Nahmias:MAC}. 

Fortunately, empirical research has demonstrated the robustness of deep learning at low bit precision. Despite being trained with 32 bit floating point representations, deep learning models can be deployed using just 4-8 bit integers without substantial accuracy degradation \cite{Jacob:quantization,  Krishnamoorthi:tf-whitepaper, banner:fourbit}. The minimum attainable precision is dependent on the network; larger neural networks, such as Resnet50, can be run at lower precision than compressed networks such as MobilenetV2 \cite{Krishnamoorthi:tf-whitepaper}.
In addition, different layers of neural networks are tolerant to different levels of bit precision, and uniformly quantizing all layers to the same low precision leads to accuracy degradation. 
Using mixed bit precision, which executes precision-sensitive layers at high precision and insensitive layers at lower precision, enables inference with as few as 2-3 bits \cite{HAWQ, HAWQv2, haq, dnas, cai:zeroq, hubara:adaquant}.
Digital hardware supports the execution of neural networks at dynamically varying bit precision; for example, NVIDIA A100 GPUs enable 1-64 bit integer arithmetic and 16-64 bit floating point arithmetic depending on the program or layer precision requirements \cite{a100}.

By contrast, analog architectures for deep learning have not exploited dynamic precision to the extent that digital architectures have. Precision in analog computers is limited by various types of noise, such as shot noise, thermal noise, and weight read noise. 
We analyze the relationship between analog noise and bit precision in Section \ref{sec:noise_bits} by equating analog noise power and quantization noise power.  We encounter a surprising observation: the effective bit precision of a fixed analog computer varies substantially for different layers,
 but is not higher for precision sensitive layers and vice versa. 
Despite this, analog matrix multipliers for deep learning do not dynamically vary precision to account for the precision sensitivity of different layers. At most, they allow for the amount of precision to be statically determined at design time for each neural network \cite{hamerly:homodyne, Nahmias:MAC, boybat:multi-memristive}, or utilize mixed precision by using a digital processor for precision-sensitive operations, such as backpropagation or the first and last layer \cite{Joshi2020AccurateDN, gallo:mpinmemory, Nandakumar:mpcm, Eleftheriou:mixed, nandakumar:mixed-analog}. 

In this work, we propose extending analog computing architectures to support dynamic precision that can be selected by a programmer or compiler, analogous to bit precision in digital hardware. 
We observe that it is possible to trade off various performance metrics, such as energy efficiency, throughput, or area, to improve the precision of the analog computing engine.
By repeating the same operation multiple times and averaging the results (as demonstrated by multi-memristive synapses \cite{boybat:multi-memristive}), precision
can be improved at the cost of expending more energy. In Section \ref{sec:mpann-hardware}, we discuss how redundant coding (repeating the same operation) in space or time can be applied to both analog electronic and optical computing architectures to enable dynamic precision.

A key challenge for deploying neural networks with dynamic precision is determining the optimal precision of different layers of the neural network given a hardware performance target. 
In Section \ref{sec:method}, we tackle this by solving an optimization problem. We focus specifically on the tradeoff between the energy per multiply-accumulate (MAC) of redundant coding and the resulting precision. We define a constrained optimization problem to maximize the original objective of the neural network subject to a constraint on total energy consumed, where the energy/MAC of each layer may be varied. The constraint is turned into a penalty function, and the optimal precision/energy tradeoff can be found by gradient descent. 

We evaluate the advantages of supporting dynamic precision in analog computing on its energy consumption via software simulations in Section \ref{sec:experiments}. We apply the method to a variety of convolutional neural networks (CNNs) for computer vision, including Resnet50, and natural language processing models such as BERT subject to different types of limiting noise, including shot noise, thermal noise, and weight noise, and determine the minimum energy/MAC with ${<}2\%$ accuracy degradation. First, we find that different networks are tolerant to different precision: some models require 34x more energy/MAC than others when each model is executed at uniform precision. 
This implies that fixed precision analog hardware will either be unable to support the networks that require greater energy/MAC, or will expend substantially more energy than needed for more noise-tolerant networks. Second, we find that using dynamic precision within a single network reduces energy consumption by up to 43-96\% while obtaining similar accuracy. Specifically, by using dynamic precision, optical homodyne photoelectric multipliers subject to the shot noise limit  \cite{hamerly:homodyne} can perform Resnet50 inference at 2.7 aJ/MAC and BERT inference at 1.6 aJ/MAC. Although this case is idealized (with no optical loss, data movement, or memory access costs included), it suggests that the energy floor for optical neural networks may be  as low as 10 aJ/MAC.
This shows the importance of designing analog hardware that supports dynamic precision: model developers will only be able to realize these performance gains if the hardware enables it.

\section{Background and Related Work}
\subsection{Deep Neural Networks}

A deep neural network consists of a series of layers, each of which performs a matrix multiplication followed by a nonlinear activation function. 
Let layer $(l)$ have $N^{(l)}$ input neurons, $N^{(l+1)}$ output neurons, weight matrix $\bfW^{(l)}$, input vector $\bfx^{(l)}$, and nonlinear activation function $f$. Then the input to the $(l+1)^{th}$ layer is computed as: 
\begin{align}
    \bfx_i^{(l+1)} = f & \left(g\left(\bfW^{(l)}_i, \bfx^{(l)}\right)\right) \quad 1 \leq i \leq N^{(l+1)} \label{eqn:nn} \\  & g\left(\bfW^{(l)}_i, \bfx^{(l)}\right) = \sum_{j=1}^{N^{(l)}}{\bfW_{ij}^{(l)}} {\bfx_{j}^{(l)}} \nonumber
\end{align} 
Each neuron, as demonstrated above, computes a dot product between a row of the weight matrix $\bfW_{i}^{(l)}$ and the input to that layer, and performs $N^{(l)}$ multiply-accumulate operations (MACs). Convolutional layers are also computed via dot products by using the $\texttt{im2col}$, or patching, method \cite{cudnn}. The neural network stacks $L$ such layers and computes the probability of class labels as $p_m(y|\bfx;\theta)$, where $\theta = \{\bfW^{(1)}, \ldots, \bfW^{(L)}\}$ denotes all the parameters of the neural network.

\subsection{Low Precision Neural Networks}
Neural networks are able to perform accurate inference at low bit precision in digital hardware. 
Empirical research has demonstrated that neural network accuracy degrades minimally when quantizing to 4 to 8-bit fixed point integer representations, despite the networks being trained using 32-bit floating point numbers \cite{Jacob:quantization, banner:fourbit, Krishnamoorthi:tf-whitepaper}. 

A common method for quantizing floating point values to low precision is affine quantization \cite{Jacob:quantization}. In affine quantization, floating point inputs $\bfx^{(l)}$ (or weights) in the range $[\bfx^{(l)}_{min},\bfx^{(l)}_{max}]$ are mapped to fixed point integers of $B$ bits from $0$ to $2^{B-1}$ by scaling, translating, and rounding the inputs. Mathematically, this is 
\begin{align}
    \bfx^{(l)}_q &=\textrm{round}\left(\frac{\bfx^{(l)}}{\Delta^{(l)}}\right)  + z^{(l)}\label{eqn:quant}\\
    &\Delta^{(l)} = \frac{\bfx^{(l)}_{max} - \bfx^{(l)}_{min}}{2^{B} - 1}\nonumber\\
    &z^{(l)} = \textrm{round}\left(\frac{- \bfx^{(l)}_{min}}{2^{B} - 1}\right) \nonumber
\end{align}
\begin{figure*}
\centering
\subfloat[Resistive Crossbar Array]{\includegraphics[width=0.25\textwidth]{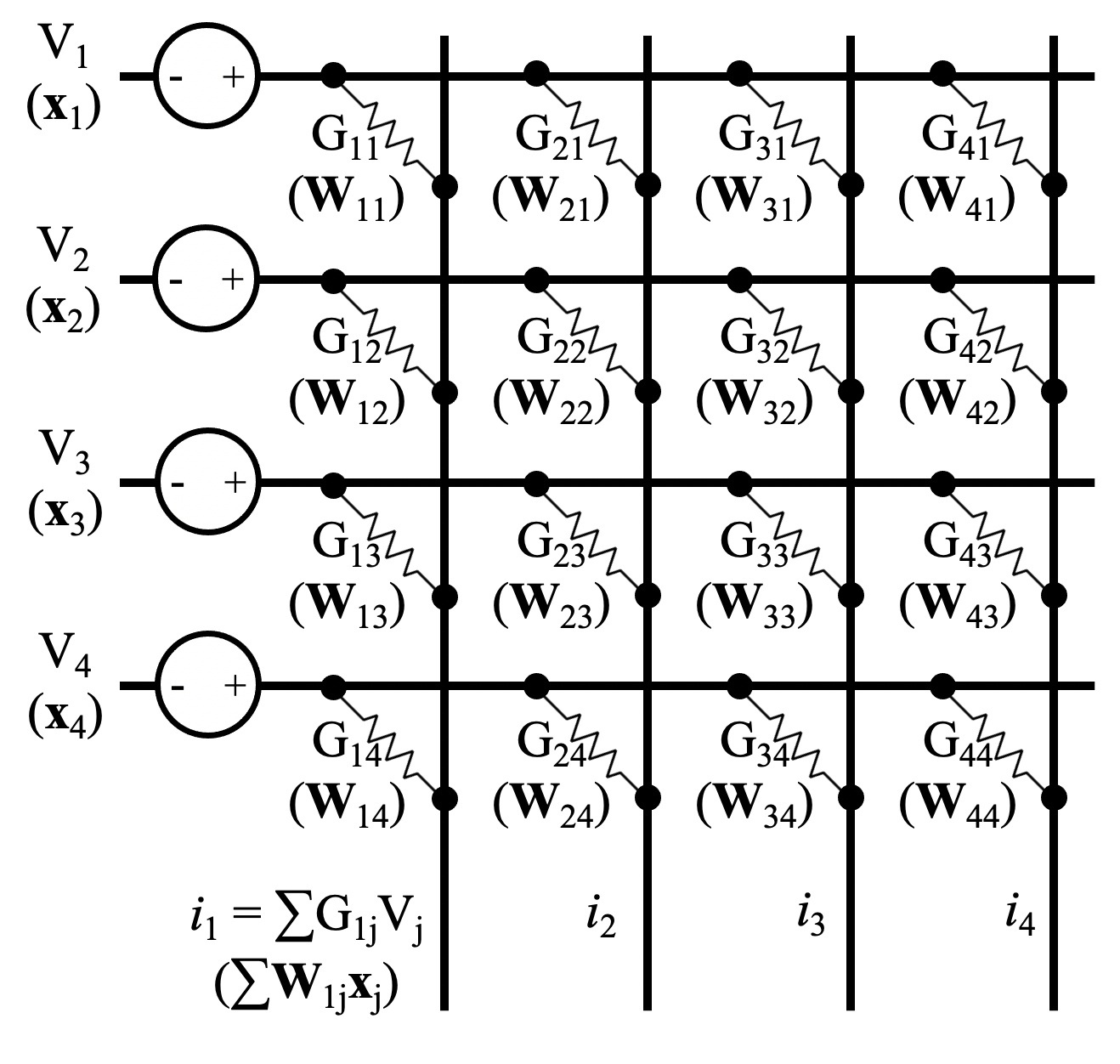}%
\label{fig:memristor}}
\hfil
\subfloat[Homodyne Photoelectric Multiplication]{\includegraphics[width=0.31\textwidth]{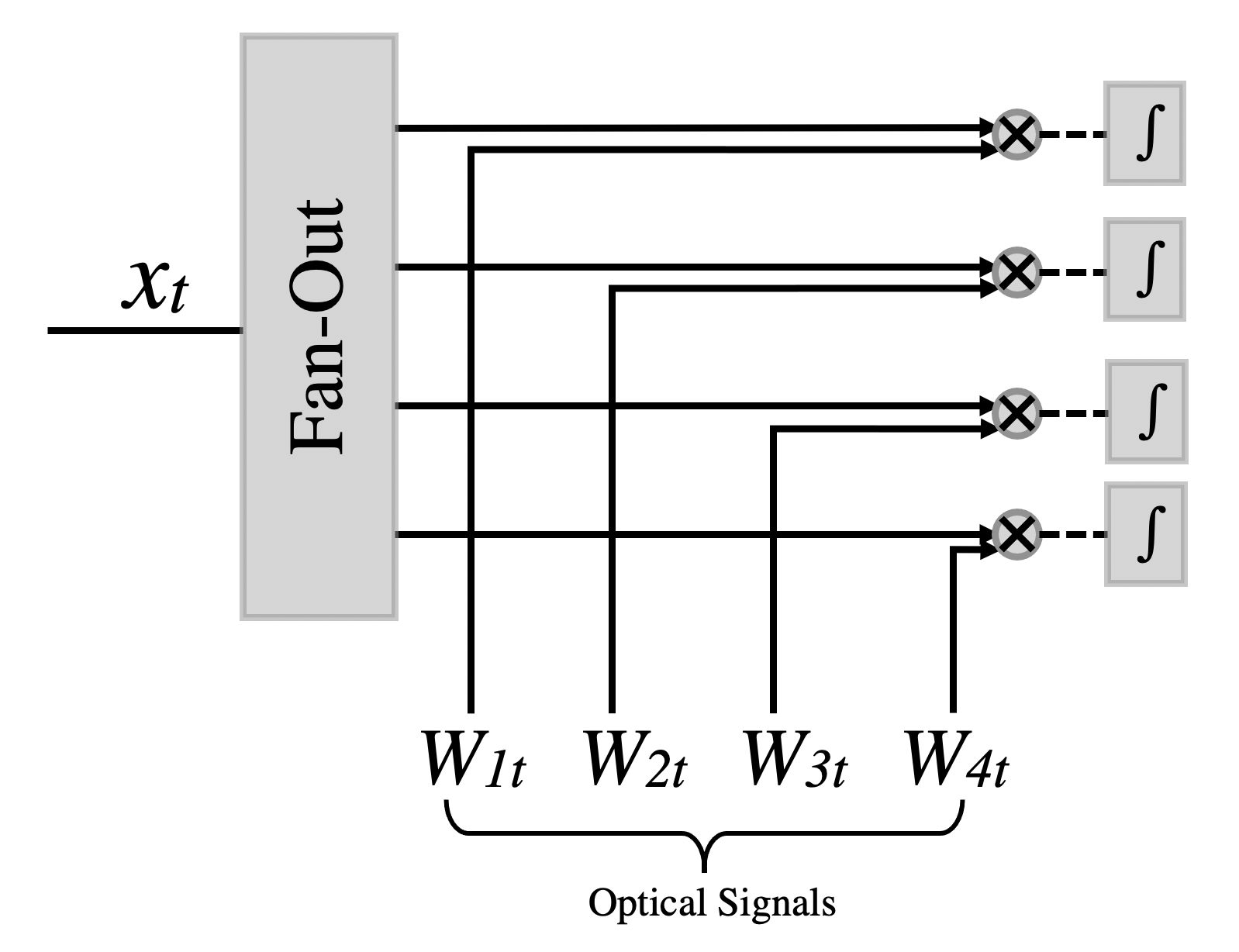}%
\label{fig:homodyne}}
\hfil
\subfloat[Broadcast and Weight]{\includegraphics[width=0.33\textwidth]{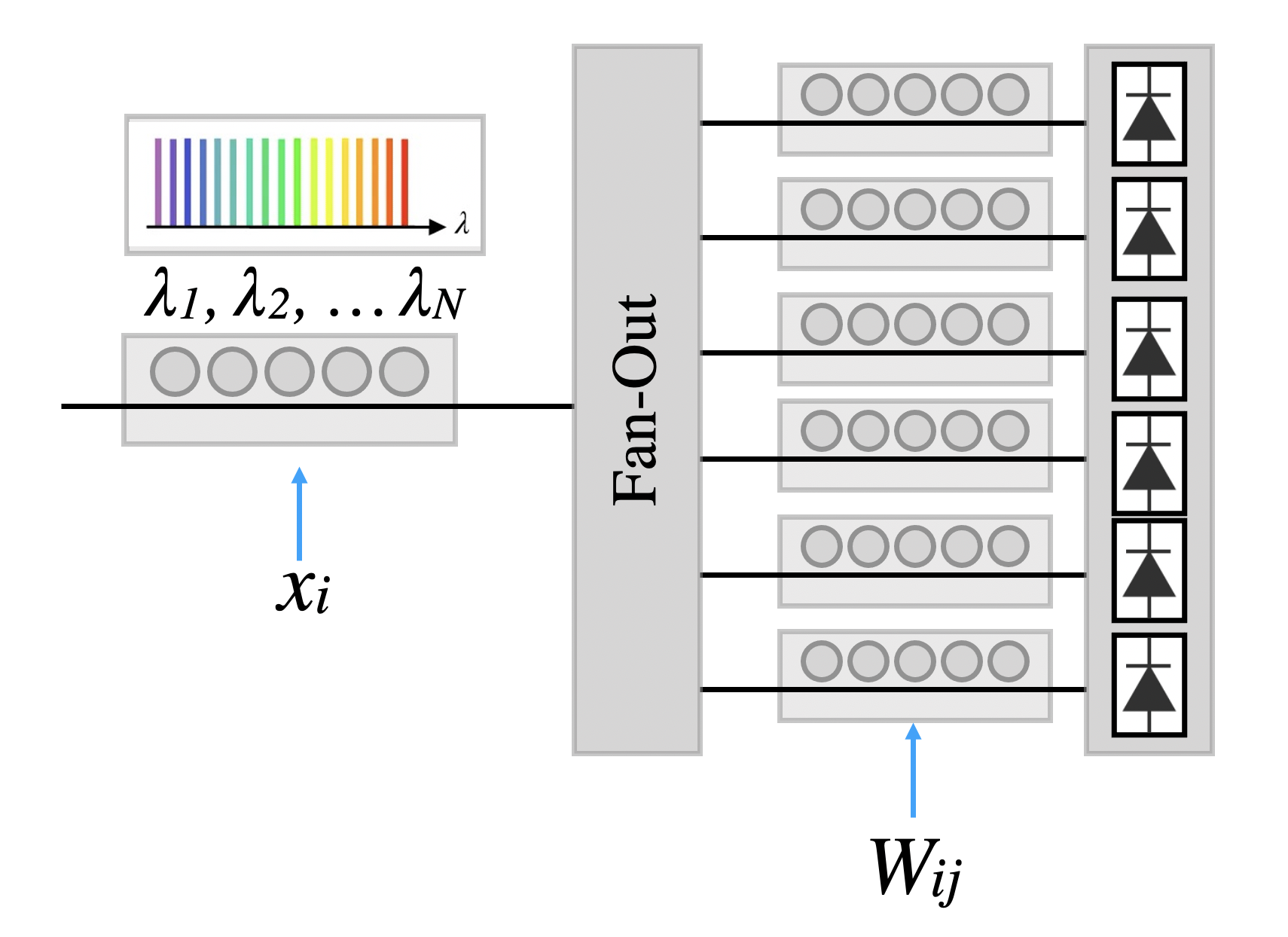}%
\label{fig:bw}}
\caption{Different approaches to analog electronic and optical computing. (a) Resistive crossbar arrays perform computing in-memory by applying inputs as voltages to rows of the array, and storing weights as conductances. Ohm's law yields a sum of products of these input voltages and weight conductances to produce a current that is proportional to the matrix-vector multiplication. (b) Homodyne photoelectric multiplication integrates charge at a coherent detector and accumulates MACs over time steps $t$. (c) Broadcast and weight uses modulators and wavelength division multiplexing to compute a matrix-vector multiplication in a single clock cycle.}
\label{fig:analog}
\end{figure*}

The average precision required by commonly deployed neural networks can be lowered by using mixed precision. Different layers of neural networks are tolerant to different degrees of precision, and uniformly quantizing all layers of a neural network to the same bit precision leads to accuracy degradation \cite{HAWQv2}. 
There are many approaches for determining the bit precision of each layer, which requires searching an exponentially large space in the number of layers \cite{HAWQ, HAWQv2, haq,dnas, banner:fourbit, cai:zeroq, hubara:adaquant}. The method most similar to the one presented in this paper learns the bitwidth of each layer via gradient descent  \cite{Uhlich:learn_bw}. Other works perform post-training mixed precision quantization by making theoretical assumptions about signal-to-quantization noise ratio (SQNR) \cite{lin:sqnr}.

\subsection{Analog Computing for Matrix Multiplication}
\label{sec:noise_background}

Analog computing improves neural network energy consumption by increasing the efficiency of linear operations such as matrix multiplications. Analog electronic computing is commonly based on resisitive crossbar arrays, which store weights in flash, memristors, or phase change memory as the conductance between two points (Figure \ref{fig:memristor}) \cite{burr:nvmem, yang:memristive_computing, tsai:analog_dl_review, li:analog_lstm, Zhongrui:analog_rl, isaac, agarwal:memristor, fick:mythic, Joshi2020AccurateDN}. Optical computing is more diverse in its architectures \cite{marinis:review, Shastri:review}, such as photoelectric multiplication in homodyne detectors (Figure \ref{fig:homodyne}) \cite{hamerly:homodyne}, broadcast-and-weight (Figure \ref{fig:bw}) \cite{Tait:Broadcast, deLima:neuromorpic-ml, peng:pics, Yang:bw, bangari:deap, tait:neuron,  tait:microring, tait:siphweight, nahmias2020laser}, and optical unitary transformations \cite{shen:coherent, Miller:15}.  

A unique characteristic of analog computing is that it is subject to noise from many different sources. These noise sources include shot noise derived from Poisson distributed photons or electron fluctuations from the incoming signal  \cite{hayat:shotnoise}, thermal noise in resistors, weight read noise in resistive memory from thermal noise, random telegraph noise, or $1/f$ noise \cite{Joshi2020AccurateDN, agarwal:requirements}, and other types of device nonlinearities or fabrication variation \cite{hu_nonlinearities}. If analog computing architectures quantize outputs to low bit precision, this noise may lead to bit errors in the least significant bit, which will not necessarily degrade neural network accuracy. 

As a result of this noise, the matrix multiplication $g$ in Equation \ref{eqn:nn} is no longer deterministic. To account for this, we replace the function $g$ with a random variable $\tilde{g}$ that is sampled based on the noise distribution. The distribution of  $\tilde{g}$ is dependent on the hardware architecture and type of noise. The noisy model's predictions are also a random variable, which we denote $\tilde{p}_m(y | \bfx; \theta)$. 

\textbf{Thermal Noise:} 
First, we consider thermal noise that occurs from a transimpedance amplifier in receiver circuitry. We consider architectures that use digital inputs and outputs.
Because the dot product is computed using quantized (i.e. normalized) versions of $\bfW$ and $\bfx$, the result $\bfW\bfx$ is recovered after rescaling the quantized product $\bfW^{(q)}\bfx^{(q)}$ by the range of $\bfW$ and $\bfx$. 
Thermal noise occurs as Gaussian noise with variance $\sigma_t^2$ (determined by receiver design) added to the quantized result, and is rescaled with the signal. 
Because longer dot products are computed via partial sums, noise variance grows linearly with the number of MACs $N^{(l)}$. So, we write:
\begin{align}
   \tilde{g}\left(\bfW^{(l)}_i, \bfx^{(l)}\right) &\sim \sum_{j=1}^{N^{(l)}}{\bfW_{ij}^{(l)}} {\bfx_{j}^{(l)}}\label{eqn:thermal} \\+ \xi \sqrt{N^{(l)}} &\left(\bfW^{(l)}_{max} - \bfW^{(l)}_{min}\right)\left(\bfx^{(l)}_{max} - \bfx^{(l)}_{min}\right)\sigma_t  \nonumber\\ 
    \xi &\sim \mathcal{N}\left(0, 1\right) \nonumber
\end{align}
These equations for thermal noise apply to both resistive crossbar arrays and optical broadcast-and-weight since both encode the signal in the current at the receiver. 

\textbf{Weight Noise:} For weight noise in resistive memory, we assume that the weights are quantized and write 
\begin{align}
   \tilde{g}\left(\bfW^{(l)}_i, \bfx^{(l)}\right) &\sim \sum_{j=1}^{N^{(l)}}{\left(\bfW_{ij}^{(l)}+ \xi_j \left(\bfW^{(l)}_{max} - \bfW^{(l)}_{min}\right) \sigma_w\right)} {\bfx_{j}^{(l)}} \nonumber \\ 
    \xi_j &\sim \mathcal{N}\left(0, 1\right) \quad  1 \leq j \leq N \label{eqn:weight}
\end{align}

\textbf{Shot Noise:} The magnitude of shot noise, however, is signal dependent. For homodyne photoelectric multipliers using analog inputs and weights, 
we have from the derivation in \cite{hamerly:homodyne}:
\begin{align}
   \tilde{g}\left(\bfW^{(l)}_i, \bfx^{(l)}\right) &\sim \sum_{j=1}^{N^{(l)}}{\bfW_{ij}^{(l)}} {\bfx_{j}^{(l)}} + \xi\frac{\norm{\bfW_i^{(l)}}_2\norm{\bfx^{(l)}}_2}{\sqrt{N^{(l)} }}\sigma_{s} \label{eqn:shot} \\ 
    \xi &\sim \mathcal{N}\left(0, 1\right) \nonumber
\end{align}

Many other systems and noise sources can be represented using this phenomenological framework. These include unitary optical matrix multiplication \cite{shen:coherent} and optical switching based architectures \cite{nahmias:switching} in both the thermal and shot noise limited regimes, resistive crossbar arrays subject to nonlinear weight noise, and optical systems with Relative Intensity Noise (RIN) \cite{Nahmias:MAC}, among others. In many of these cases, the noise is not linear, such as when noise is added to power, but signals are encoded in amplitudes \cite{shen:coherent}.

While prior works in analog computing discuss the varying noise tolerance of different computations and propose using mixed-precision, to our knowledge, this is the first work that extends analog computing to support programmatically dynamic precision. Prior work statically increases the precision for a single neural network by using greater energy/MAC \cite{boybat:multi-memristive, hamerly:homodyne, Nahmias:MAC}, simulates hardware that injects noise into only a single layer of the neural network \cite{hamerly:homodyne}, or uses mixed analog and digital computing, where precision-sensitive operations such as the first and last layer \cite{Joshi2020AccurateDN}, backpropagation \cite{nandakumar:mixed-analog}, and others \cite{Eleftheriou:mixed, Nandakumar:mpcm, gallo:mpinmemory} are computed digitally  at high precision while other operations are computed subject to analog noise. Other approaches include using arithmetic codes to correct errors in analog computing \cite{feinberg:arithmetic}. Recent work in stochastic computing has introduced dynamic precision to improve efficiency \cite{sim:stochastic}.

\section{Relating Noise and Bit Precision} 
\label{sec:noise_bits}
To better understand noise-limited precision in analog computing, we establish the relationship between analog precision and bit precision. We do so by treating quantization to low bit precision as an additive uniform noise source \cite{sripad:qerror, gray:qerror, baskin:nice} and evaluating the number of bits for which the variance of analog noise is equal to the variance of quantization noise. We define analog noise in each layer with a scalar random variable $\epsilon_a^{(l+1)}$. We define the noise distribution over the entire layer, as opposed to each neuron, because quantization to low bit precision for activations is performed at a per-layer granularity. 

We begin by evaluating the variance of the quantization noise, $\epsilon_q$. When using uniform quantization, all activations in layer $(l+1)$ are quantized with $B$ bits in the range $\left[\bfx^{(l+1)}_{max}, \bfx^{(l+1)}_{min}\right]$, so each quantization bin has width $\Delta^{(l+1)} = \left(\bfx^{(l+1)}_{max} - \bfx^{(l+1)}_{min}\right) / (2^B - 1)$. We model quantization as uniform noise added in the interval $[-{\Delta^{(l+1)}}/{2}, {\Delta^{(l+1)}}/{2}]$, which has variance $\Delta^2/12$. So, the variance of quantization noise is 
\begin{align}
    \textrm{Var}(\epsilon_q^{(l+1)}) =  \frac{1}{12}\left(\frac{\bfx^{(l+1)}_{max} - \bfx^{(l+1)}_{min}}{2^{B} - 1}\right)^2
\end{align}
While the number of bits used for quantization is discrete, we may treat the variance of quantization noise as a continuous function of a fractional number of bits.

Now, we can define the number of bits of noise precision, or noise bits $B_\epsilon$, as the number of bits for which $\textrm{Var}(\epsilon_a^{(l+1)}) = \textrm{Var}(\epsilon_q^{(l+1)})$.  Solving the equation for $B_\epsilon$, we get 
\begin{align}
    B_\epsilon^{(l+1)} =\log_2{\left({\frac{\bfx^{(l+1)}_{max} - \bfx^{(l+1)}_{min}}{\sqrt{12\textrm{Var}\left(\epsilon_a^{(l+1)}\right)}}} + 1\right)} \label{eqn:noise_bits}
\end{align}
Note that $ B_\epsilon^{(l+1)}$ is not an information-theoretic quantity. While we could compute the number of noise bits for a layer as the mutual information of $g, \tilde{g}$, this does not capture the relationship with bit precision using a \textit{uniform} quantizer, which is typically the case in digital hardware for efficiently implementing MACs. 

We observe a connection between Equation \ref{eqn:noise_bits} and signal-to-noise ratio (SNR). We define $\textrm{SNR}^{(l+1)} = {\textrm{Var}(S^{(l+1)})}/{\textrm{Var}(\epsilon_a^{(l+1)})}$, where the signal in layer $(l+1)$ is $S^{(l+1)}$. 
If signals in layer $(l+1)$ were distributed uniformly over their range, then $\textrm{Var}(S^{(l+1)}) = \left(\bfx^{(l+1)}_{max} - \bfx^{(l+1)}_{min}\right)^2/12$, and we would have $B_\epsilon^{(l+1)} = \log_2\left(\sqrt{\textrm{SNR}^{(l+1)}} + 1 \right)$. However, prior works observe that the signal distribution of neural networks is not uniform, and often has has much smaller variance and quantization efficiency than a uniform distribution over the same range \cite{lin:sqnr,sheng:rangecomp}. This is different from a classical communication channel, where the signal distribution is typically assumed to be fixed and uniform \cite{you:coding}. 

Finally, we characterize the number of noise bits explicitly for thermal noise using the variance in Equation \ref{eqn:thermal}: 
\begin{align}
    &B_\epsilon^{(l+1)} = \label{eqn:range-comp} \\&\log_2{\left({\frac{\left(\bfx^{(l+1)}_{max} - \bfx^{(l+1)}_{min}\right)}{\sigma_t\left(\bfW^{(l)}_{max} - \bfW^{(l)}_{min}\right)\left(\bfx^{(l)}_{max} - \bfx^{(l)}_{min}\right)\sqrt{12N^{(l)}}}} + 1\right)}\nonumber
\end{align}

We evaluate whether noise bits accurately capture precision by testing whether they  predict the inference accuracy of neural networks subject to analog noise.
To do so, we evaluate each layer of Resnet50 subject to thermal noise with varying $\sigma_t$, and measure the number of noise bits in each layer. We then remove thermal noise, and instead run each layer at low bit precision using its respective number of noise bits.\footnote{To more closely capture the continuous valued number of noise bits, we perform quantization to a fractional number of bits by rounding up the number of bins. For example, quantization over 25 uniformly spaced bins requires 4.644 bits.} We report results in Table \ref{tab:noise_bits_before}, including the average number of noise bits across all layers. 
We find that the accuracy subject to analog noise and quantization error closely match one another, despite the distributions of noise being different. The quality of the approximation decreases at extremely low precision.
    
\begin{figure}
\centering
\includegraphics[width=0.475\textwidth]{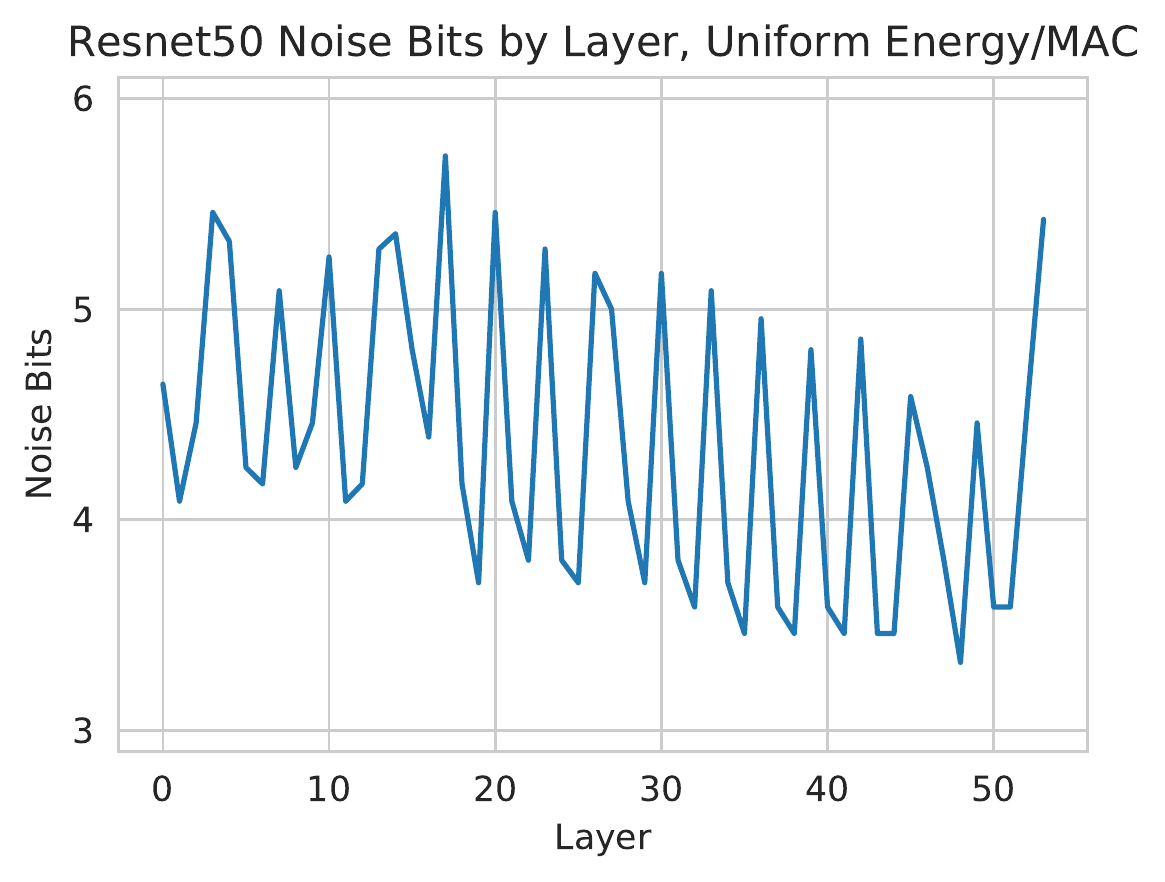}%
\label{fig:thermal}
\caption{Number of bits of noise precision when using fixed $\sigma_t$ for different layers of Resnet50.}
\label{fig:thermal_noise_bits_before}
\end{figure}

\begin{table}[]
\centering
\begin{tabular}{|p{0.06\textwidth}|p{0.1\textwidth}|p{0.06\textwidth}|p{0.12\textwidth}|}
\hline
Noise $(1000\sigma_t)$ & Noisy Accuracy & Average Bits $B_\epsilon$ & Low Bit Accuracy \\ \hline
7.1 & 36.5 & 3.2 & 39.1 \\ \hline
4.5 & 66.4 & 3.8 & 61.7 \\ \hline
3.2 & 71.2 & 4.3 & 70.0 \\ \hline
2.2 & 73.2 & 4.8 & 73.1 \\ \hline
1.8 & 73.8 & 5.1 & 73.9 \\ \hline
1.6 & 74.1 & 5.3 & 74.2 \\ \hline
1.4 & 74.4 & 5.4 & 74.5 \\ \hline
1.0 & 74.9 & 5.9 & 75.0 \\ \hline
0.7 & 75.3 & 6.4 & 75.2 \\ \hline
0.4 & 75.4 & 7.1 & 75.4 \\ \hline
0.0                      & 75.5           & 8.0                       & 75.5             \\ \hline
\end{tabular}
\vspace{2mm}
\caption{Thermal Noise and Equivalent Bit Precision for Resnet50}
\label{tab:noise_bits_before}
\end{table}

We plot the number of noise bits for each layer of Resnet50 in Figure \ref{fig:thermal_noise_bits_before}. We note a surprising observation: even when using a fixed $\sigma_t$, the number of noise bits in different layers varies substantially. This is because the number of noise bits is a function of the dynamic range compression from inputs to outputs of layer $(l)$, as explicitly described in Equation \ref{eqn:range-comp}.

While the number of noise bits varies for different layers, there is no guarantee that precision sensitive layers are executed at high precision and vice versa. A heuristic evaluation demonstrates this is not the case: the first and last layer typically require the highest precision \cite{Joshi2020AccurateDN}, but are not allocated higher precision in Figure \ref{fig:thermal_noise_bits_before}. 
To address this problem, we propose a method to vary the precision settings of analog computers in Section \ref{sec:mpann-hardware} and a method to determine the optimal precision for different layers in Section \ref{sec:method}.

\section{Dynamic Precision with Redundant Coding}
\label{sec:mpann-hardware}
\begin{figure*}[t!]
\centering
\subfloat[Time Averaging]{\includegraphics[width=0.265\textwidth]{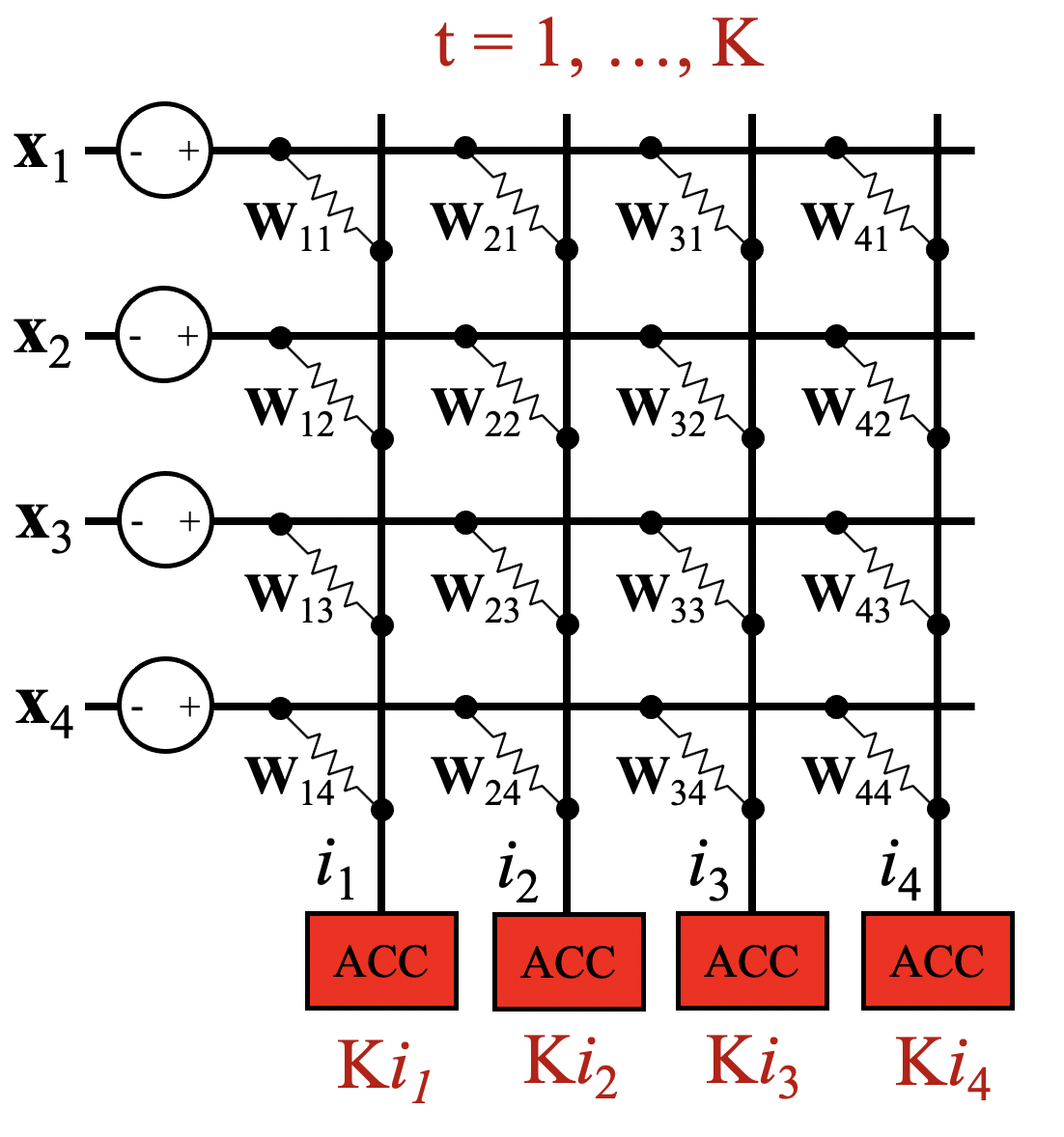}%
\label{fig:time}}
\hfil
\subfloat[Spatial Averaging per Matrix]{\includegraphics[width=0.35\textwidth]{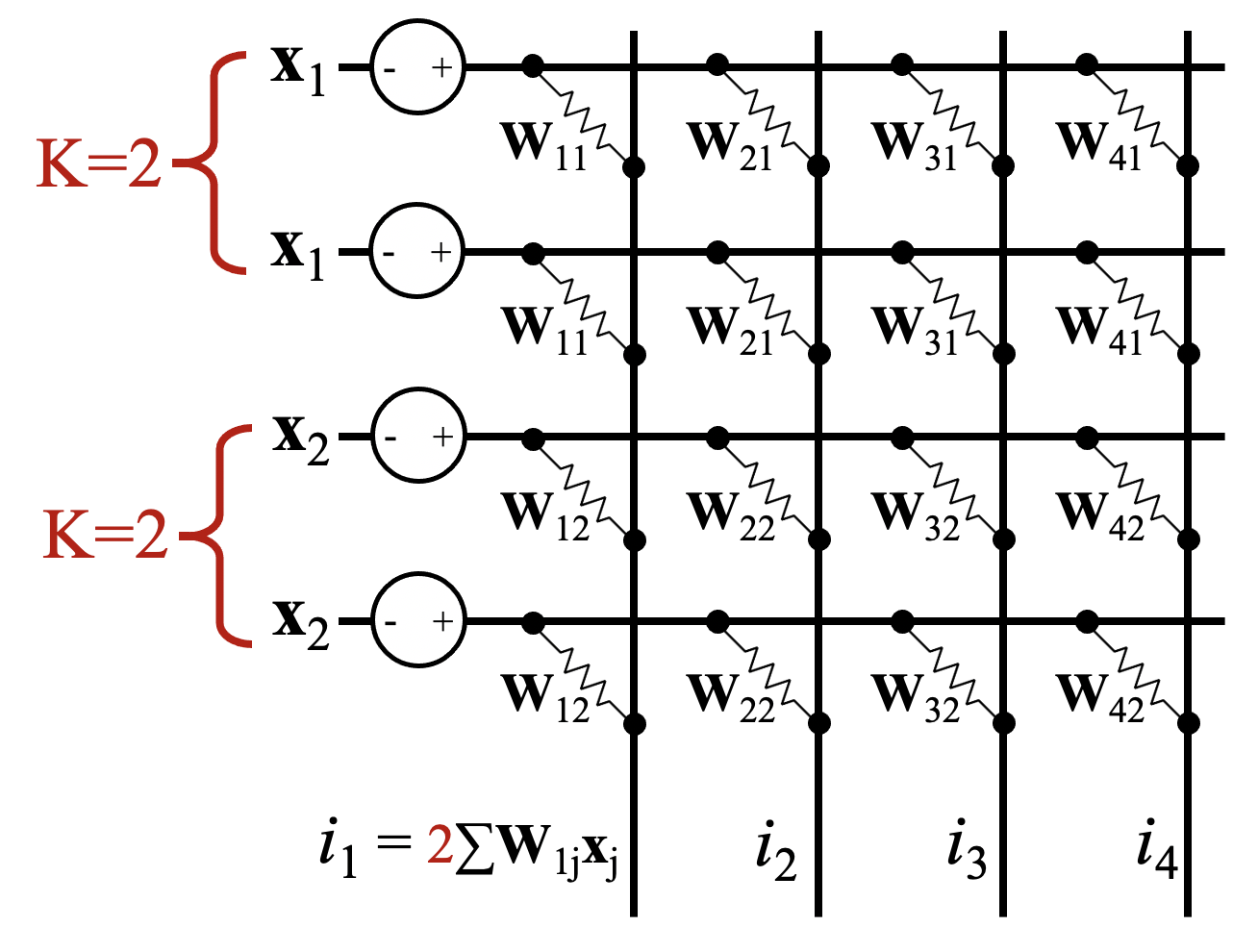}%
\label{fig:spatial}}
\hfil
\subfloat[Spatial Averaging per Dot Product]{\includegraphics[width=0.27\textwidth]{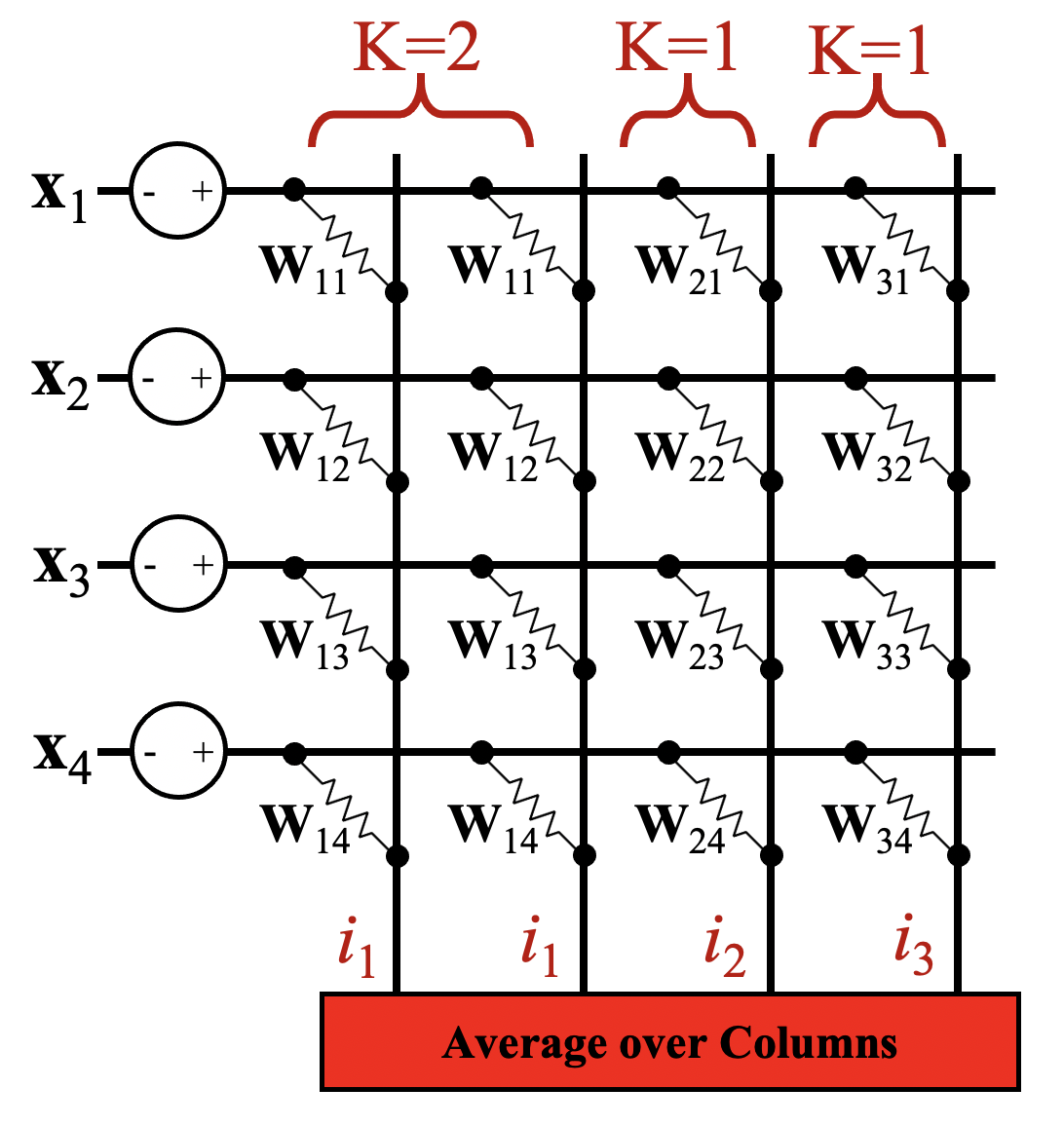}%
\label{fig:channel}}
\caption{Dynamic precision with redundant coding; resistive crossbar arrays are used as an illustrative example. Changes to the architecture are shown in red. We use $K$ to denote the number of times an operation is repeated, where in (a) operations are repeated for $K$ clock cycles, in (b) the same inputs and weights are repeated, and in (c) only certain rows of $\bfW$ are repeated. }
\label{fig:redundant-coding}
\end{figure*}

We propose extending analog computing architectures to support dynamic precision through a general method called redundant coding. Redundant coding entails performing the same computation multiple times, either in different spatial channels of the analog matrix-vector multiplier, or over multiple clock cycles, and averaging the result. This reduces the impact of noise on the computation at the expense of other performance metrics, such as energy/MAC, throughput, or compute density. Redundant coding has previously been demonstrated as a method for improving the precision of analog computation by using multiple memristors to encode the same weight \cite{boybat:multi-memristive}. 
This work generalizes redundant coding as a technique applicable to all analog electronic and optical computing architectures, and proposes designing architectures that can programmatically vary the amount of redundancy, as opposed to statically improving the precision with a fixed amount of redundancy.

We first demonstrate how redundant coding can be used to vary precision at the granularity of a matrix multiplication. To enable time averaging, receiver circuitry may add an accumulator, and the compiler can instruct the hardware to accumulate the same computation for $K$ clock cycles and average the result before requantizing. To enable spatial averaging, $K$ devices may be used to encode the same weights and inputs in a single dot product. For example, in resistive crossbar arrays, multiple resistive memory elements in a column 
can be used to encode the same weight, and the same input voltage can be broadcasted to multiple rows. The broadcasting of weights and inputs to multiple devices can be determined at compile time. The application of redundant coding to a resisitve crossbar array via time and spatial averaging are shown in Figures \ref{fig:time} and \ref{fig:spatial}. With $K$ times redundant coding, both of these approaches effectively compute 
\begin{align}
K\bfW\bfx = \begin{bmatrix}
\bfW & \bfW & \ldots & \bfW
\end{bmatrix} \begin{bmatrix}
\bfx\\
\bfx\\
\vdots\\
\bfx
\end{bmatrix} \nonumber
\end{align}

Dynamic precision at the finer granularity of each row of the weight matrix requires modifying spatial averaging to use a varying number of dot product engines to repeat different computations. Then, the architecture will need to average over a programmable number of dot product engines. This can either be performed by configuring receiver circuitry to support an accumulator over variable numbers of engines, or by leveraging a digital ALU and designing instructions for performing this averaging. The configuration of these accumulators or the instructions for averaging can be determined at compile time. We illustrate spatial averaging for dot products in a resisitve crossbar array in Figure \ref{fig:channel}. Similar approaches have been used by resisitve crossbar arrays to split the computation of different bits over different columns \cite{isaac}; here, we propose that the accelerator do so dynamically.

The precision of the computation varies with the degree of redundancy. Because the amount of redundancy linearly increases the energy/MAC, we parameterize precision with respect to $E^{(l)}$, the amount of energy/MAC used for the $(l)^{th}$ layer of the neural network. We consider the ideal case where this quantity may be continuously modulated, as opposed to taking one of several quantized energy levels. In each of the cases discussed in Section \ref{sec:noise_background}, the signals add linearly, but the noise sources add in quadrature, so the noise standard deviation is proportional to $\frac{1}{\sqrt{E^{(l)}}}$. We note that the type of averaging also determines other tradeoffs that are made; time averaging trades off throughput, and spatial averaging trades off area.

\textbf{Thermal Noise:} We replace Equation \ref{eqn:thermal} with: 
\begin{align}
   & \tilde{g}\left(\bfW^{(l)}_i, \bfx^{(l)}, E^{(l)}\right) \sim \sum_{j=1}^{N^{(l)}}{\bfW_{ij}^{(l)}} {\bfx_{j}^{(l)}}  \nonumber \label{eqn:thermal-vary} \\ & +  \xi \sqrt{N^{(l)}} \left(\bfW^{(l)}_{max} - \bfW^{(l)}_{min}\right)\left(\bfx^{(l)}_{max} - \bfx^{(l)}_{min}\right) \frac{\sigma_t}{
  \sqrt{E^{(l)}}}
   \end{align}
   
\textbf{Weight Noise:} We replace Equation \ref{eqn:weight} with:
\begin{align}
   &\tilde{g}\left(\bfW^{(l)}_i, \bfx^{(l)}, E^{(l)}\right) \nonumber\\ & \sim  \sum_{j=1}^{N^{(l)}}{\left(\bfW_{ij}^{(l)}+ \xi_j \left(\bfW^{(l)}_{max} - \bfW^{(l)}_{min}\right)\frac{\sigma_w}{\sqrt{E^{(l)}}}\right)} {\bfx_{j}^{(l)}} \label{eqn:weight-vary} 
\end{align}
For thermal noise and weight noise, $E^{(l)}$ is a relative and unitless quantity because the free parameters $\sigma_t$ and $\sigma_w$ are determined by the engineering of a given architecture.

\textbf{Shot Noise:} For shot noise, we may modify Equation \ref{eqn:shot} to have $E^{(l)}$ represent a physical, not relative, energy quantity for specific architectures such as \cite{hamerly:homodyne}. In this case, $E^{(l)}$ is measured in Joules, $E^{(l)}\lambda / (hc)$ is the average number of photons per MAC, and the output subject to shot noise in homodyne photoelectric multipliers \cite{hamerly:homodyne} is: 
\begin{align}
   \tilde{g}\left(\bfW^{(l)}_i, \bfx^{(l)}, E^{(l)}\right) &\sim \sum_{j=1}^{N^{(l)}}{\bfW_{ij}^{(l)}} {\bfx_{j}^{(l)}}  \label{eqn:shot-vary} \\ &+ \xi\frac{\norm{\bfW_i^{(l)}}_2\norm{\bfx^{(l)}}_2}{\sqrt{N^{(l)}E^{(l)}\lambda / (hc) }} \nonumber
\end{align}

\section{Learning Optimal Precision-Energy Tradeoffs}
\label{sec:method}

A key challenge for deploying neural networks with dynamic precision is determining the optimal precision of different layers of the neural network given a hardware performance target. In this work, we focus primarily on the tradeoff between energy/MAC and precision resulting from redundant coding. To do so, we propose to solve a constrained optimization problem to maximize the original objective of the neural network subject to an energy constraint. Note that this optimization problem is solved for a pretrained network and only optimizes over the energy allocated to each layer, so it does not retrain the neural network. We use $E_{max}$ to denote the energy budget, and $\bfE$ to denote the vector of all energies to be learned across layers $1, \ldots, L$, i.e. $(E^{(1)}, \ldots, E^{(L)})$. The total energy consumed by the network can be computed from the number of MACs in each layer, $n_{mac}^{(l)}$, as $E_{tot}(\bfE) = \sum_{l=1}^LE^{(l)}n_{mac}^{(l)} $. The objective of the neural network, in this case log likelihood, is evaluated on ordered pairs of inputs (i.e. images) and outputs (i.e. classification labels) $(\bfx, y)$ sampled from the data distribution $p_{d}(\bfx, y)$.  Then, the optimization problem is 
\begin{equation}
\begin{aligned}
\minimize_{\bfE} \quad & -\mathbb{E}_{(\bfx, y) \sim p_{d}}\left[\log \tilde{p}_m(y | \bfx; \theta, \bfE)\right]\\
\textrm{s.t.} \quad & \sum_{l=1}^LE^{(l)}n_{mac}^{(l)} \leq E_{max}\\
\end{aligned}
\end{equation}

We address the fact that $\tilde{p}_m$ is a random variable by using the reparameterization trick \cite{kingma:vae}. We can treat the noise $\xi^{(l)}_i \sim \mathcal{N}(0, 1)$ as inputs to the network, in which case $\tilde{p}_m$ becomes a deterministic function of the noise, weights, and inputs. Let $\xi$ denote the random vector of all noise sources. The new optimization problem is 
\begin{equation}
\begin{aligned}
\minimize_{\bfE} \quad & -\mathbb{E}_{(\bfx, y) \sim p_{d}, \xi^{(l, i)} \sim \mathcal{N}(0, 1)}\left[\log \tilde{p}_m(y | x, \xi; \theta,  \bfE)\right]\\
\textrm{s.t.} \quad &  \sum_{l=1}^LE^{(l)}n_{mac}^{(l)} \leq E_{max}\nonumber\\
\end{aligned}
\end{equation}

We turn the linear constraint into a penalty in the objective via the Lagrange multiplier penalty method \cite{lagrange}: 
\begin{equation}
\begin{aligned}
\minimize_{\bfE} \quad & -\mathbb{E}_{(\bfx, y) \sim p_{d}, \xi}\left[\log \tilde{p}_m(y | x, \xi; \theta, \bfE)\right]  \\&+ \lambda \max\left(\sum_{l=1}^LE^{(l)}n_{mac}^{(l)} - E_{max}, 0\right)  
\end{aligned}
\end{equation}
where $\lambda \in \mathcal{R}^{+}$ is a fixed hyperparameter used for weighting for the penalty term. Solving this optimization problem does not guarantee that the constraint is fulfilled. As the hyperparameter $\lambda$ is increased, the weighting of the penalty dominates the loss and the constraint is more likely to be fulfilled.

This objective can now directly be optimized with respect to the energy per layer via stochastic gradient descent. To train the energy allocations, we can use the Monte Carlo estimator of the objective by sampling data and noise. Because this minimization is performed only over $\bfE$, not the parameters $\theta$, the optimization problem can typically be solved with a small number of gradient steps on a small subset of the original training dataset.  

Finally, we find that in practice, penalizing the logarithm of total energy consumption (an equivalent quantity) is more stable for optimization. This is because it is advantageous to have energy allocations and the log likelihood to be similar orders of magnitude in the loss, but the energy allocations change by orders of magnitude during training. This yields our final optimization problem, 
\begin{equation}
\begin{aligned}
& \minimize_{\bfE} \quad  -\mathbb{E}_{(\bfx, y) \sim p_{d}, \xi}\left[\log \tilde{p}_m(y | x, \xi; \theta, \bfE)\right]  \\&\quad \quad + \lambda \max\left(\log\left(\sum_{l=1}^LE^{(l)}n_{mac}^{(l)}\right) - \log\left(E_{max}\right), 0\right)  
\end{aligned}
\label{eqn:logopt}
\end{equation}

Energy can also be allocated to computations at a finer grained scale than each layer, such as for each channel of a convolutional neural network or each row of a weight matrix. Since each weight channel is convolved over the entire input image, it is reasonable to have dynamic precision by channel, as is done in digital approaches \cite{banner:fourbit}. In this case, we learn $E^{(l,i)}$, or the energy/MAC for the $i^{th}$ channel in layer $(l)$. 

One challenge for the aforementioned method is if the function $\tilde{g}$ includes quantization operations, such as rounding. The gradient of the round function is zero almost everywhere, so the method will not naively be able to learn energy allocations by gradient descent. Following the literature for quantization aware training, we use the Straight Through Estimator (STE) to resolve this issue, effectively computing $\textrm{grad}_x \left(\textrm{round}(x)\right) = 1$ \cite{bengio:ste}. Then, dynamic precision can be learned in the presence of deterministic quantization.

While the problem above assumes that energy/MAC is continuous, this method can also be applied when restricted to quantized energy levels, as in the case of redundant coding. This can be done by rounding the energy/MAC to the nearest quantized energy level during training using the STE.

\section{Experiments}
\begin{table*}[t!]
\centering
\begin{tabular}{|l|l|l|l|l|l|l|}
\cline{3-7}
      \multicolumn{2}{c|}{}           & Resnet50 & Mobilenet & Inceptionv3 & Googlenet & Shufflenetv2 \\ \cline{3-7} \hline
\multirow{4}{2.7cm}{Shot Noise Energy/MAC (aJ)} 
& Uniform           &  25.4    &  62.3   &    39.0    &   31.5   &    72.0    \\  \cline{2-7}
&Dynamic Per Layer   & 4.5     & 21.2     &    8.7    &  10.4    &    24.0     \\ \cline{2-7}
&Dynamic Per Channel & 2.8    &  15.3    &     4.8   &   6.5    &   18.3      \\ \cline{2-7}
&Improvement       &  89.0\%      &  75.4\%    &    87.7\%     &   79.4\%   &    74.6 \%     \\\hline \hline
\multirow{4}{2.7cm}{Thermal Noise  Energy/MAC (relative)} 
&Uniform           &  36.5    & 2812    &    52.3    &   47.8   &    1369     \\ \cline{2-7}
&Dynamic Per Layer   &  13.5    &  333.5   &   29.7     &   27.7   &    182.2     \\ \cline{2-7}
&Dynamic Per Channel &   8.1   &  122.0   &   16.1     &    18.1   &    110.7     \\ \cline{2-7}
&Improvement       &   77.8\%   &   95.7\%   &   69.2\%       &     62.1\%  &    91.9\%    \\ \hline \hline 
\multirow{4}{2.7cm}{Weight Noise Energy/MAC (relative)} 
&Uniform           &  131.0    &  1027   &    316.2    &    198.1  &   500.0      \\ \cline{2-7}
&Dynamic Per Layer   &  48.8    &  296.2   & 149.2    &    115.5  &    295.1     \\ \cline{2-7}
&Dynamic Per Channel &   37.2   &  263.1    &  122.5 &    113.7   &    241.3     \\ \cline{2-7}
&Improvement       &   71.6\%     &  74.4\% &    61.3\%     &   42.6\%     &  51.7\%       \\ \hline
\end{tabular}
\vspace{2mm}
\caption{Minimum Energy/MAC  with ${<}2\%$ Accuracy Degradation}
\label{tab:results}
\end{table*}        

\label{sec:experiments}
\subsection{Setup}

We evaluate the impacts of dynamic precision on computer vision models and natural language processing models. For computer vision models, we evaluate five image classification models, Resnet50 \cite{he:resnet}, Mobilenetv2 \cite{sandler:mobilenet}, Inceptionv3 \cite{szegedy:inception}, Googlenet \cite{Szegedy:googlenet}, and Shufflenetv2 \cite{ma:shufflenet}, on the ImageNet dataset \cite{deng:imagenet}. For natural language models, we evaluate BERT, a popular transformer architecture  \cite{bert}, fine-tuned for the GLUE MNLI entailment task \cite{glue}. All results are reported on the corresponding validation datasets. We train optimal energy allocations $\bfE$ using 4\% of the training dataset for one epoch, which takes ${<}10$ minutes on a NVIDIA V100 GPU, a small fraction of the time to train the entire model. Unless otherwise noted, we assume energy/MAC may be continuously varied.
For each task, we determine the minimum average energy/MAC for which the accuracy does not degrade below floating point accuracy by 2\% (within 0.1\%) by performing a binary search on the target energy/MAC. 

We evaluate computer vision models subject to three different noise sources: shot noise, thermal noise, and weight noise, using Equations \ref{eqn:thermal-vary}-\ref{eqn:shot-vary}. For shot noise applied to homodyne photoelectric multipliers, we report absolute optical energy consumption in aJ, using a photon energy of 128zJ at $\lambda=1.55\mu m$ and a photodetector responsivity of $\rho=1$ \cite{hamerly:homodyne}. Inputs and weights are continuous-valued, as in neuromorphic computing. 
For thermal noise and weight noise, which are dependent on architectural implementation and engineering details captured by the parameters $\sigma_t, \sigma_w$, we report the energy/MAC as a relative, unitless quantity. Inputs and weights are digital, using 8-bit uniform quantization. Quantization parameters are calibrated on a small subset of the training data  \cite{Jacob:quantization}. Because thermal noise variance is dependent on the range of each layer, we clip activations at the 99.99th percentile, following \cite{li:percentileclip, mckinstry:percentile}. Additional setup details are in Appendix \ref{app:details}, and additional experimental results, such as on the impact of percentile clipping, are in Appendix \ref{app:more_results}. Code is open-sourced at \url{https://github.com/sahajgarg/low_precision_nn}.

\subsection{Results}

\begin{figure}[t!]
\centering
\includegraphics[width=0.475\textwidth]{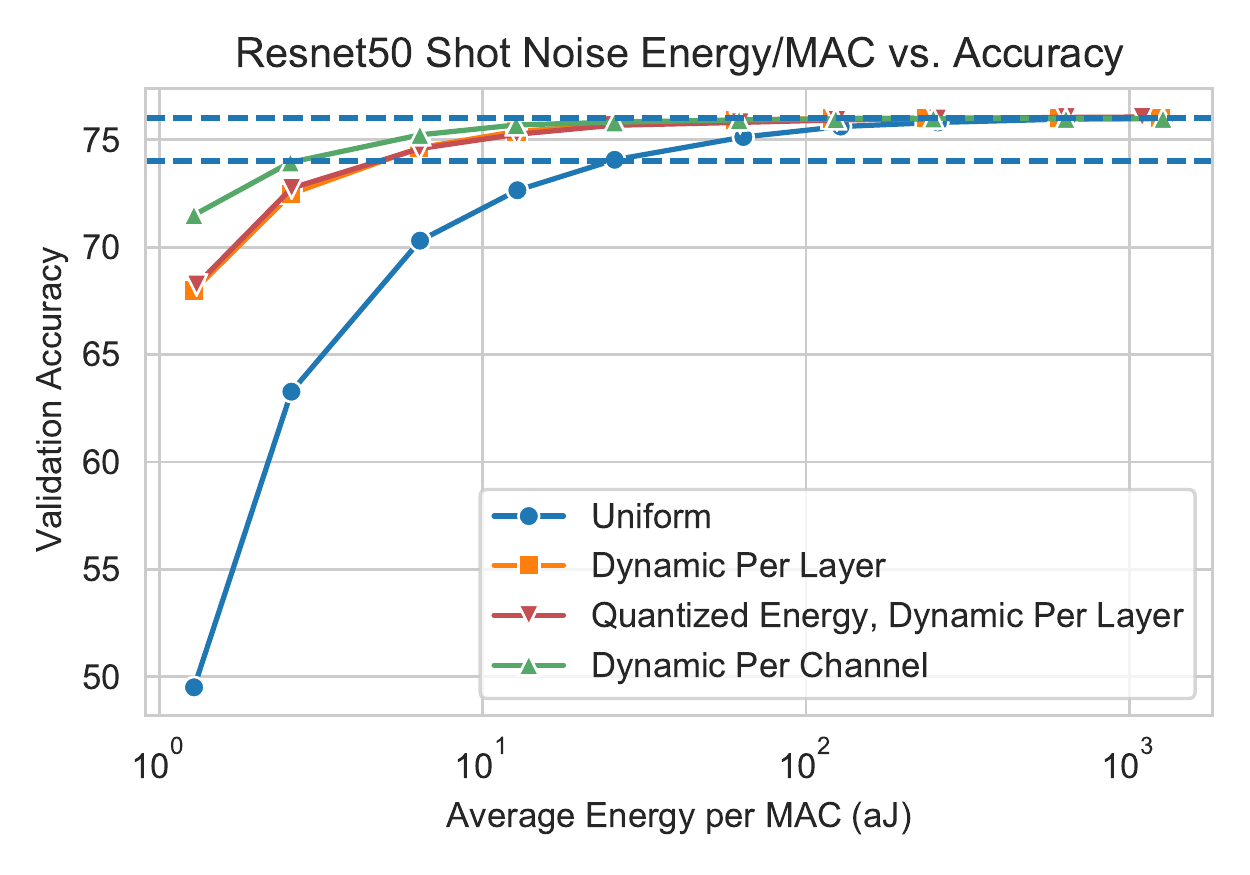}
\caption{Accuracy improves with energy/MAC, which reduces the impact of noise. Utilizing dynamic precision improves the allocation of energy to different layers and consequently inference accuracy.}
\label{fig:results}
\end{figure}

We report the minimum attainable energy/MAC with ${<}2\%$ accuracy degradation for computer vision models when using uniform precision, dynamic precision per layer, and dynamic precision per channel in Table \ref{tab:results}. 
We find that using dynamic precision within a neural network can reduce energy consumption of a single model by 43-96\%. The largest improvement is observed for MobilenetV2 subject to thermal noise because Mobilenet is especially sensitive to lowered bit precision \cite{Krishnamoorthi:tf-whitepaper}, and thus requires substantially higher energy/MAC for certain layers. Results are consistent across all three noise sources. We show the tradeoff between optical energy/MAC and accuracy in Figure \ref{fig:results}, including when energy/MAC for each layer is constrained to be a discrete number of photons. The constraint on quantized energy levels does not noticeably affect results. 

These results suggest that larger computer vision models may actually be more computationally efficient than smaller computer vision models.
Resnet50 obtains the lowest energy/MAC in Table \ref{tab:results}, likely because it is the largest of the five models in terms of number of MACs, and consequently is the most overparameterized. We evaluate the total energy consumption of  Resnet50 and MobilenetV2 when using dynamic precision for Resnet50 to obtain the same accuracy as MobilenetV2. In this case, Resnet50 requires just 0.997 aJ/MAC optical energy consumption, and despite performing 13.6x more MACs than MobilenetV2, consumes 11\% less total 
optical energy. As suggested in \cite{hamerly:homodyne}, this reinforces the importance of designing energy efficient architectures, and not necessarily compressed or small architectures.   

\begin{figure}
\centering
\includegraphics[width=0.475\textwidth]{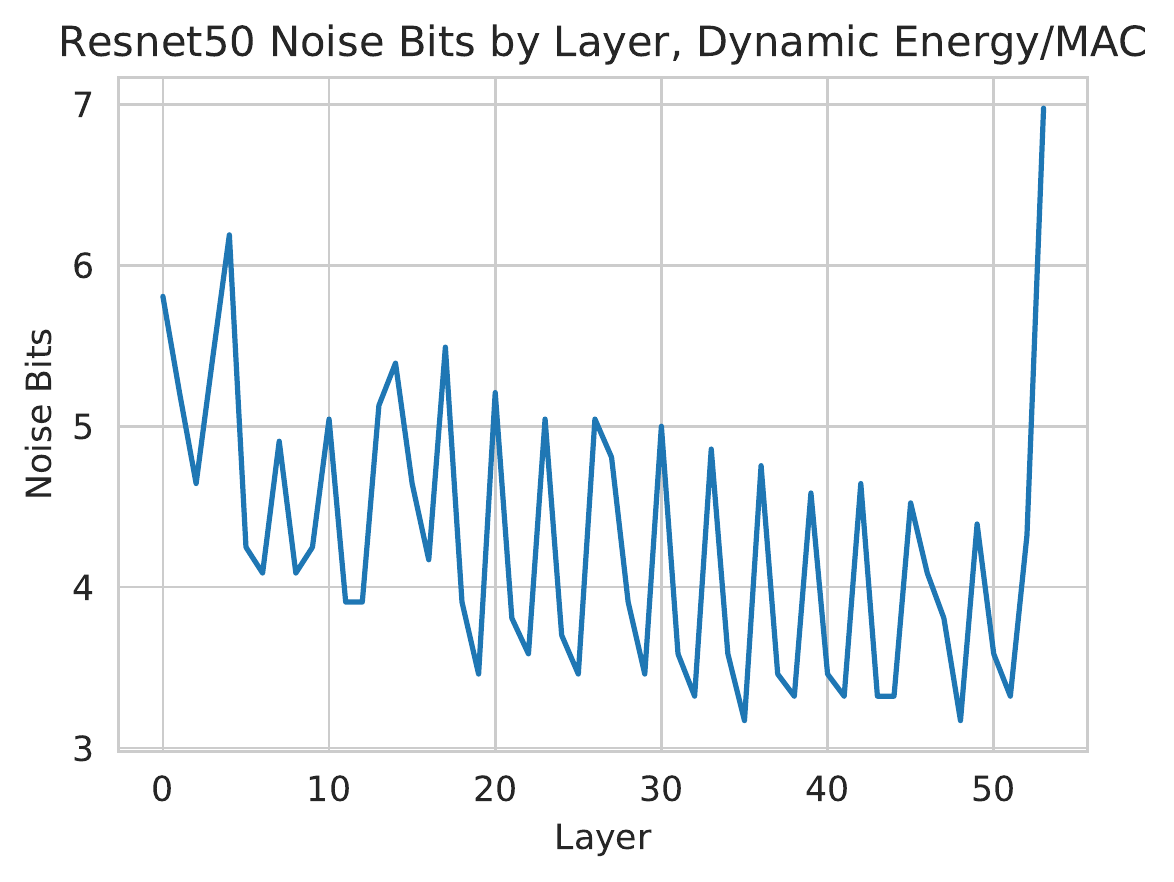}%
\label{fig:thermal}
\caption{Number of noise equivalent bits of precision when using dynamic energy/MAC for different layers of Resnet50.}
\label{fig:thermal_noise_bits_after}
\end{figure}

We show that the relationship between analog noise and noise bits still holds when using dynamic energy/MAC in Table \ref{tab:noise_bits_after}. Comparing the rows in Table \ref{tab:noise_bits_before} and Table \ref{tab:noise_bits_after}, which correspond to the same average energy/MAC, we observe that the average number of noise bits for uniform and dynamic precision is similar, but the accuracy of the dynamic precision model is higher because it more effectively allocates noise bits to precision-sensitive layers. This demonstrates why we define noise bits per layer. We show the noise bits per layer when using dynamic precision in Figure \ref{fig:thermal_noise_bits_after}, and find that the first several and last layer are executed at higher effective bit precision, unlike when using fixed energy/MAC in Figure \ref{fig:thermal_noise_bits_before}. 

\begin{table}[]
\centering
\begin{tabular}{|p{0.1\textwidth}|p{0.1\textwidth}|p{0.06\textwidth}|p{0.12\textwidth}|}
\hline
Average Energy/MAC & Noisy Accuracy & Average Bits $B_\epsilon$ & Low Bit Accuracy \\ \hline
2   & 51.1 & 3.2 & 43.2 \\ \hline
5   & 70.8 & 3.8 & 68.3 \\ \hline
10  & 73.7 & 4.3 & 73.0 \\ \hline
20  & 74.7 & 4.8 & 74.3 \\ \hline
29  & 74.9 & 5.1 & 74.7 \\ \hline
39  & 75.0 & 5.3 & 75.0 \\ \hline
50  & 75.2 & 5.5 & 75.1 \\ \hline
99  & 75.3 & 6.0 & 75.2 \\ \hline
196 & 75.4 & 6.4 & 75.3 \\ \hline
488 & 75.4 & 7.1 & 75.4 \\ \hline
\end{tabular}
\vspace{2mm}
\caption{Dynamic Precision with Thermal Noise and Equivalent Bit Precision for Resnet50}
\label{tab:noise_bits_after}
\end{table}

\begin{figure}[t!]
\centering
\includegraphics[width=0.475\textwidth]{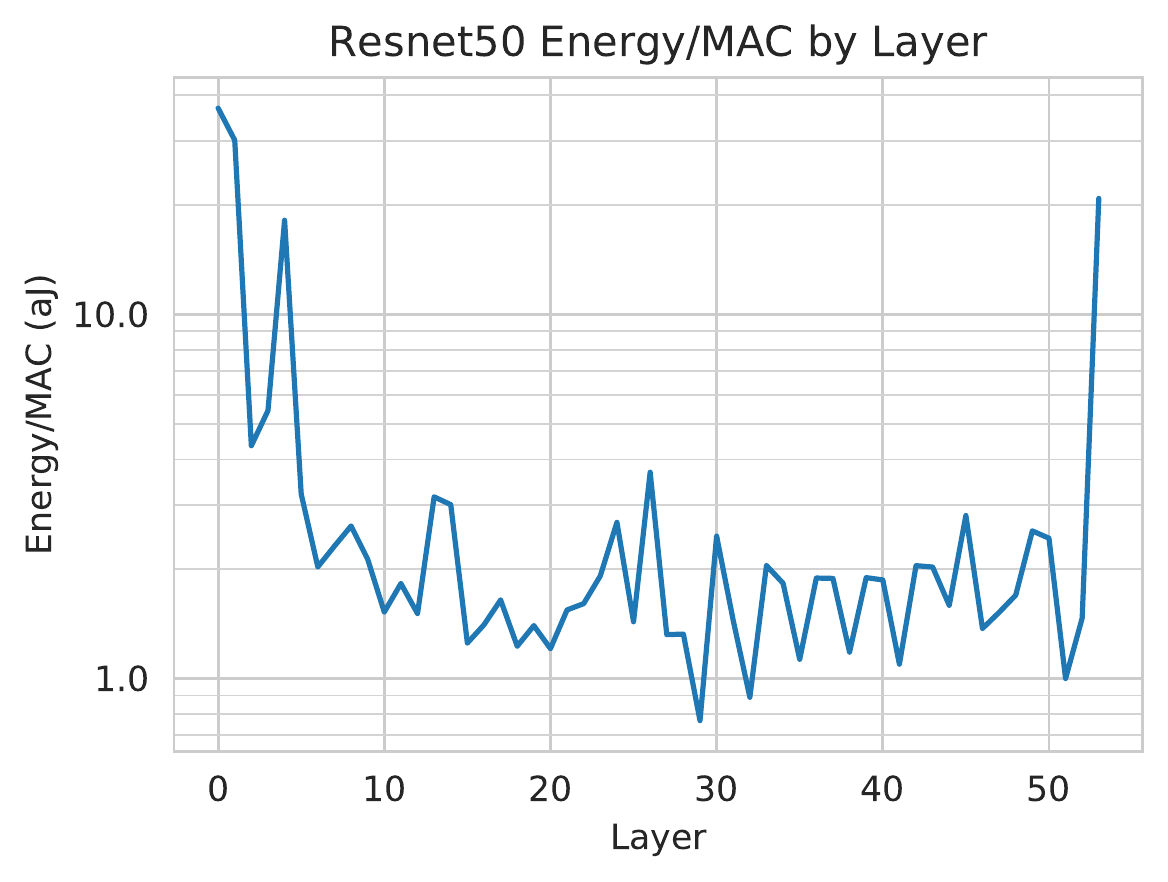}
\caption{Allocations of energy to each layer are complex. The first several and last layer are allocated higher energy/MAC, and the allocations follow sawtooth patterns due to the 3-layer building blocks of the network architecture.}
\label{fig:powers}
\end{figure}

We examine the dynamic energy allocations per layer of Resnet50 in Figure \ref{fig:powers} to better understand why dynamic precision improves performance. The energy/MAC varies substantially by layer: the first few and last layers require ${>}10$x the energy/MAC of other layers. We infer that if these layers are not run at high precision, the neural network will not produce accurate results. Hence, using uniform precision per layer will require using the energy/MAC needed for the most sensitive layer. We further observe that the final energy allocations are complex. This is similar to the observations on mixed precision inference for digital neural networks \cite{Uhlich:learn_bw, haq, dnas, HAWQv2}. These observations emphasize the importance of using an empirical, automatic method instead of manually or analytically determining the required precision.

Finally, we evaluate the minimum energy/MAC required for BERT inference on the MNLI Dataset in Table \ref{tab:bert}. Using dynamic precision per layer improves BERT energy consumption by 24\% to just 1.6 aJ/MAC, lower than any of the computer vision models. The energy improvement for using dynamic precision in BERT is smaller than for computer vision models. This is likely because no large layers in BERT require high precision, unlike the first and last layer of computer vision models, which are precision sensitive and perform around $20\%$ of the total MACs in the network. The energy/MAC for different matrix multiplications in BERT is reported in Appendix \ref{app:more_results}. Regardless, dynamic precision in analog hardware is necessary to enable BERT inference at low energy/MAC while also allowing for the higher energy settings required by computer vision models. 

\begin{table}[t!]
\centering
\begin{tabular}{|l|l|}
\hline
              & MNLI Dataset \\ \hline 
 Uniform           & 2.1
 \\  \hline
Dynamic Per Layer   & 1.6
\\ \hline
Improvement   & 24\%
\\ \hline
\end{tabular}
\vspace{2mm}
\caption{BERT Shot Noise Constrained Energy/MAC (aJ)}
\label{tab:bert}
\end{table}

\section{Discussion}
These results emphasize the importance of designing analog computing architectures that can support programmable and dynamic precision. The required energy/MAC when using uniform precision for different models can range from 2.1 aJ/MAC for BERT to 72 aJ/MAC for ShufflenetV2. If analog architectures do not support dynamic precision, then new neural networks may require higher precision than is supported, and render existing hardware nonfunctional. Moreover, enabling dynamic precision allows programmers to develop novel techniques for lowering the energy requirements of analog neural networks. This work demonstrates one such technique for utilizing dynamic precision for different layers or channels of neural networks to reduce energy consumption by 43-96\%. Moreover, to our knowledge, this is the first work that examines large transformer models such as BERT subject to analog noise.

By enabling dynamic precision, this work demonstrates that the bound on optical energy consumption of homodyne photoelectric multipliers, set by shot noise in photodetectors, can be as low as 1.6 aJ/MAC. This extends the result in \cite{hamerly:homodyne} for MLPs and shallower convolutional networks like AlexNet to deeper models such as Resnet50, for which the bound on energy/MAC is as low as 2.7 aJ/MAC. We note that this energy consumption is for an idealized system with no optical loss in the shot noise limited regime, and does not account for energy consumption of data movement, analog to digital conversion, or memory traffic. These figures are primarily meant as a demonstration that dynamic precision can be used to prevent optical matrix multiplication energy expenditure from being the system bottleneck.

In addition to varying the precision in analog computing resulting from noise, it is possible to vary the bit precision of analog architectures that use digital inputs, weights, and/or outputs. When using dynamic precision, many inputs and weights may be subject to noise of sufficiently large magnitude that several of the least significant bits are discarded. We see this in Figure \ref{fig:thermal_noise_bits_after}, where some layers of Resnet50 use fewer than 4 noise bits but are quantized to 8 bit integers. It may be possible to dynamically set analog-to-digital converter precision based on the number of bits of noise precision in different computations. 
Moreover, the optimization problem in Equation \ref{eqn:logopt} can be extended to jointly learn the optimal number of bits to allocate per layer, as done by \cite{Uhlich:learn_bw} for digital architectures. 

Depending on the hardware architecture, the total energy penalty may need to be modified. In this work, we only model the energy consumed by the matrix multiplier; however, substantial energy is consumed by data movement, portions of which which may be done digitally, or by other operations such as partial sum accumulation or nonlinearities \cite{Nahmias:MAC}. Approaches such as time averaging only require moving more data if the bit precision of inputs is increased, so the energy/MAC may not scale linearly with the amount of redundancy if data movement is modeled. If the bit precision of different layers is also learned, the energy penalty may also include memory pressure based on the bitwidth of activations and weights, analog-to-digital converter energy consumption, and other factors. 

Finally, even more energy efficient models may be enabled by training neural networks to be more noise tolerant. Many approaches to obtaining low bit precision require retraining the network parameters while simulating the effect of quantization to obtain high accuracy \cite{Krishnamoorthi:tf-whitepaper, HAWQv2, Uhlich:learn_bw}. A similar noise-aware training process can be applied to jointly learning network parameters and dynamic precision allocations by modifying the optimization problem in Equation \ref{eqn:logopt} to also optimize over the parameters $\theta$ of the neural network and other quantizer parameters such as the range of different layers (as in  \cite{esser:lsq}), which affect noise magnitude. Such approaches may enable sub-aJ/MAC neural network inference. 

\section{Conclusion}
In this work, we demonstrate the utility of extending analog computing architectures to support dynamic precision with redundant coding. By repeating operations and averaging the result, redundant coding enables programmable tradeoffs between precision and other desirable performance metrics, such as energy efficiency or throughput. Enabling dynamic precision is critical for supporting different models that require different precision: for example, Shufflenetv2 requires 3x the energy/MAC of Resnet50, and 34x the energy/MAC of BERT. Moreover, we show that it is possible to leverage dynamic precision within a single model by solving an optimization problem to maximize log likelihood while adding a penalty for energy consumption, and that using dynamic precision within a model improves energy consumption by 43-96\%. In one example of optical neural networks limited by shot noise, dynamic precision enables Resnet50 inference at an optical energy consumption of just 2.7 aJ/MAC and BERT at 1.6 aJ/MAC with ${<}2\%$ accuracy degradation. 
These results emphasize the importance of designing analog architectures to support dynamic precision.


%
\appendices
\section{Experimental Details}
\label{app:details}

We provide additional details for the experimental setup. Scripts for generating results in the paper are released at \url{https://github.com/sahajgarg/low_precision_nn}.

\textbf{Noise}: For thermal noise, we set $\sigma_t = 0.01$ and for weight noise, we set $\sigma_w = 0.1$. These choices were arbitrary, so energy/MAC is referred to as a relative quantity in the main text. For specific architectures, they should be measured. We assume that residual connections, concatenation operations, max pooling, and average pooling occur without additional noise. We restrict our evaluation for BERT to shot noise because self-attention layers of BERT, which multiply two activation matrices, may be challenging to compute in-memory. 

\textbf{Quantization}: We use per-channel quantization of weights and per-tensor quantization of activations \cite{Krishnamoorthi:tf-whitepaper}. For linear layers, we perform quantization per row of the weight matrix, analogous to per-channel quantization of convolutional layers. Weight quantization parameters are determined by calibration on the min/max values of the weights in each channel. When evaluating subject to thermal noise, the minimum and maximum values for activations are set based on the 99.99th percentile of the data, evaluated over 120 training examples \cite{li:percentileclip, mckinstry:percentile}. Because percentile clipping degrades accuracy by 0.3\% when activations are at high precision, but improves accuracy at lower activation precision (shown in Appendix \ref{app:more_results}), it is used only for thermal noise. A better strategy for clipping parameters may be to learn the quantizer minimum and maximum, as in \cite{esser:lsq}, jointly with energy allocations, which we leave to future work. For weight noise, activations are calibrated based on a moving average of the min/max values of the data over 100 batches with a batch size of 32. The skip and residual connections are quantized, and the outputs are requantized to 8 bits. 

\textbf{Training}: Energy allocations are trained using the Adam optimizer with a learning rate of 0.01 \cite{kingma:adam}. The penalty hyperparameter was set to $\lambda=2$ for shot noise, and $\lambda=8$ for thermal and weight noise; we found that the method was relatively insensitive to the choice of $\lambda$. We did not do extensive experimentation with respect to the required dataset size, but found that the method was relatively insensitive to the use of less calibration data than presented in this work, assuming the energy allocations are trained until convergence. 

\textbf{Evaluation}: Accuracy degradation of all models except Mobilenet is measured with respect to the respective floating point baseline. For Mobilenet, 8 bit quantization degrades accuracy by ${>}1\%$, so we evaluate 2\% accuracy degradation relative to the 8 bit baseline. For BERT evaluation, the MNLI entailment task reports both matched and mismatched accuracy on entailment. We measure the average accuracy degradation of the two metrics. 

\section{Additional Results}
\label{app:more_results}

\begin{figure}[t!]
\centering
\includegraphics[width=0.475\textwidth]{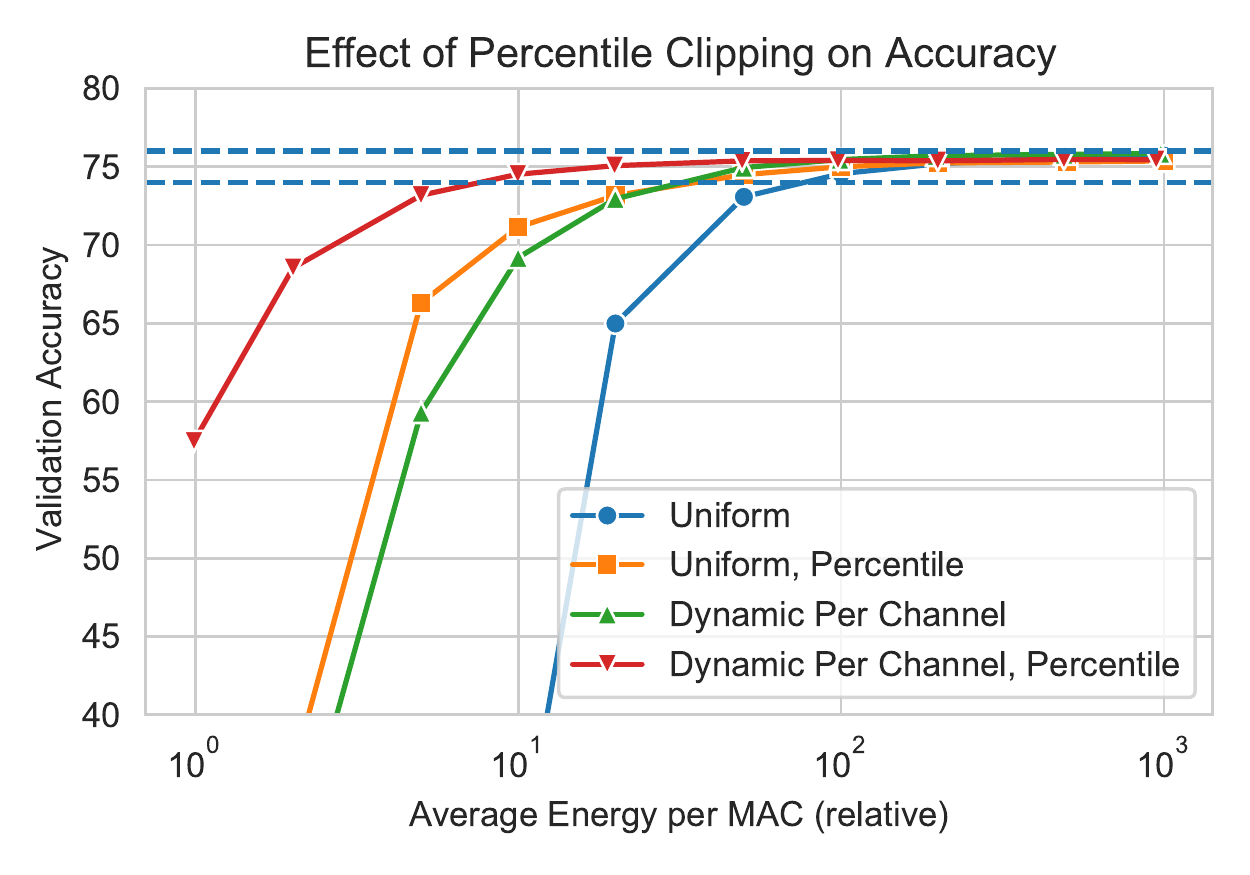}
\caption{Percentile clipping of activations improves accuracy subject to thermal noise.}
\label{fig:percentile}
\end{figure}

In Figure \ref{fig:percentile}, we evaluate the impact of percentile clipping on accuracy subject to thermal noise. We find that with high amounts of noise, percentile clipping of activations improves accuracy of both uniform precision and dynamic precision models, and uniform precision with percentile clipping outperforms dynamic precision without clipping. This is likely because the magnitude of thermal noise is proportional to the range of inputs, and clipping at the 99.99th percentile reduces the range by approximately half. However, at high precision, it degrades accuracy by 0.3\%. 

\begin{figure}[t!]  
\centering
\includegraphics[width=0.45\textwidth]{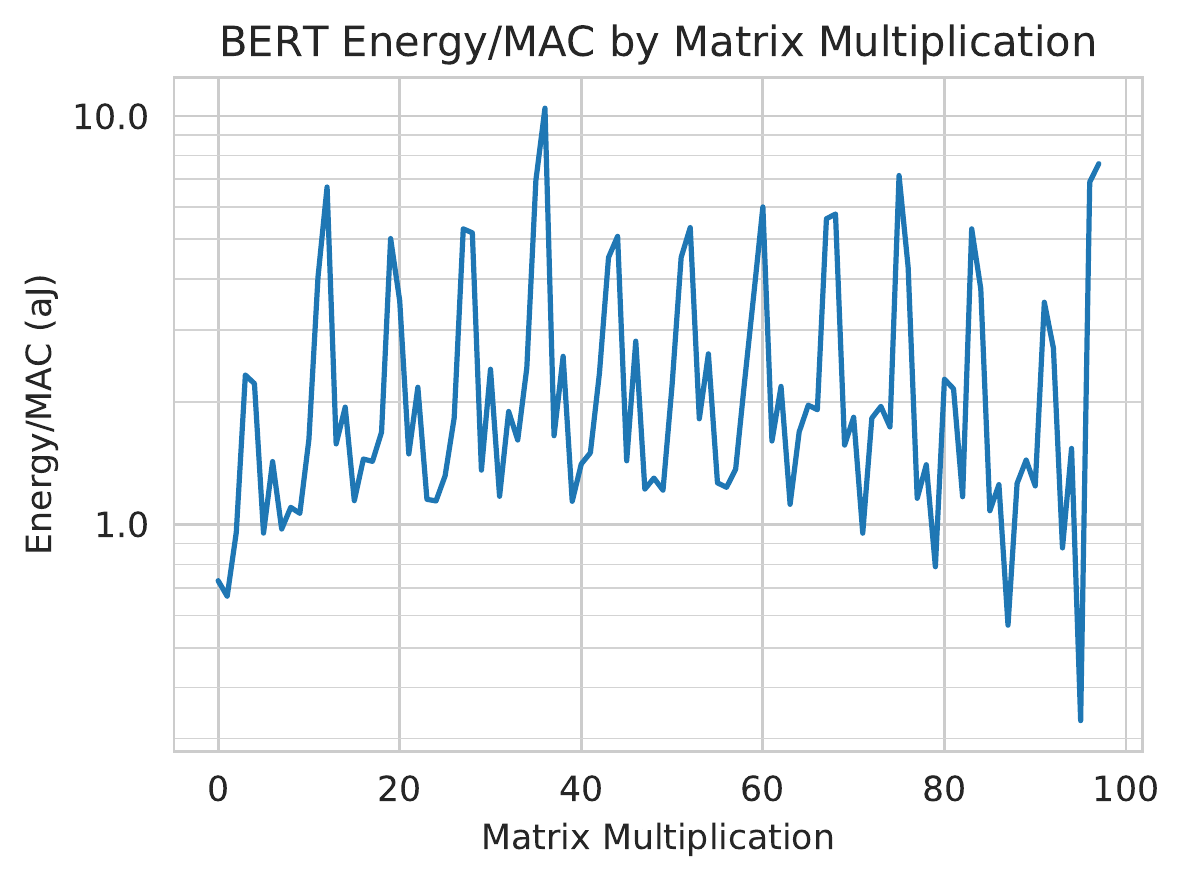}
\caption{Bert Energy/MAC for different matrix multiplications in BERT. Note that each layer performs multiple matrix multiplications. While some layers utilize more energy/MAC, such as the last layer, these layers perform an extremely small fraction of the total MACs in the network and consequently do not contribute substantially to the total energy consumption.}
\label{fig:bert_power}
\end{figure}

\begin{figure}[t!]
\centering
\includegraphics[width=0.45\textwidth]{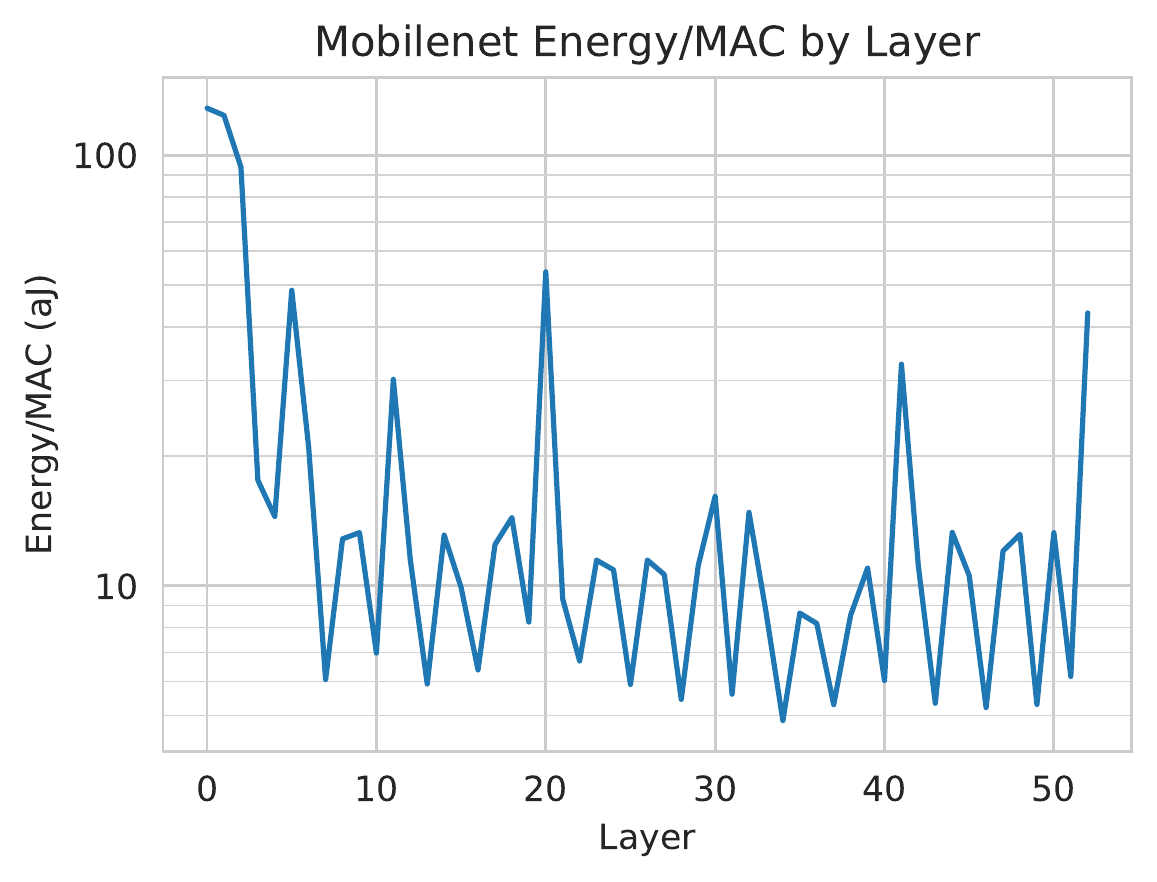}
\caption{Allocations of energy to each layer for MobilenetV2 demonstrate similar results to Resnet50.}
\label{fig:powers_mobilenet}
\end{figure}

In Figure \ref{fig:bert_power}, we show the varying energy/MAC for different matrix multiplications in BERT, and in Figure \ref{fig:powers_mobilenet}. we show the varying energy/MAC for different layers of MobilenetV2. 


\newpage

\section*{Acknowledgment}
We would like to thank many members of the Luminous Computing team, including Michael Gao, Matthew Chang, Rodolfo Camacho-Aguilera, Rohun Saxena, Katherine Roelofs, and Patrick Gallagher for their helpful discussions and suggestions.


\ifCLASSOPTIONcaptionsoff
  \newpage
\fi



%

\bibliographystyle{IEEEtran}
\bibliography{IEEEabrv,bibliography.bib}




%









\end{document}